# Automating lookahead planning using site appearance and space utilization

Eyob Mengiste [a,b], Borja García de Soto [a], Timo Hartmann [b]

[a] S.M.A.R.T. Construction Research Group, Division of Engineering, New York University Abu Dhabi (NYUAD), Experimental Research Building, Saadiyat Island, P.O. Box 129188, Abu Dhabi, United Arab Emirates

[b] Technical University of Berlin, Berlin, Germany

Email: eyob.mengiste@nyu.edu, garcia.de.soto@nyu.edu, timo.hartmann@tu-berlin.de

**Abstract**

This study proposes a method to automate the development of lookahead planning. The proposed method uses construction material conditions (i.e., appearances) and site space utilization to predict tasks' completion rates. A Gated Recurrent Unit (GRU) based Recurrent Neural Network (RNN) model was trained using a segment of a construction project's timeline to estimate completion rates of tasks and propose data-aware lookahead plans. The proposed method was evaluated in a sample construction project involving finishing works such as plastering, painting, and installing electrical fixtures. The results show that the proposed method can assist with developing automated lookahead plans. In doing so, this study links construction planning with actual events at the construction site. It extends the traditional scheduling techniques and integrates a broader spectrum of site spatial constraints into lookahead planning.

***Keywords:*** *Recurrent Neural Network, Gated Recurrent Unit, Look Ahead Planning; Lean Construction; LPS; Reality Capture*

## 1   Introduction

The construction industry is trying to enhance productivity, efficiency, and reliability by employing new technologies, methodologies and tools. In practice, construction projects are complex and have multiple stakeholders, generally resulting in delays and cost overruns (Al Hosani et al., 2020). The necessity to ensure seamless execution has led to the development of concepts like Lean Construction (Howell, 1999), with its foundational principles focusing on minimizing waste and improving overall project outcomes.

The Lean Construction Institute's Last Planner System ® (LPS) (Ballard, 2020) is one such pivotal innovation that seeks to reduce lead times, costs, and other forms of waste through effective communication and coordination among all parties involved. LPS has five stages (Heigermoser et al., 2019), and Lookahead planning is one of them. Lookahead planning requires accurate information and effective collaboration, which has been identified as one of its significant challenges. In particular, Lookahead planning often lacks complete and timely information from stakeholders, thus leading to limited utilization (Perez and Ghosh, 2018). The value of precise data in lookahead planning cannot be overstated, and an emerging solution to improve the accuracy and reliability of this data is the integration of real-time data from construction sites into project schedules. The traditional practice of construction scheduling often defines tasks in a way that is disconnected from the evolution of the construction site. This disconnect can lead to oversight of the on-ground realities of the construction project.

Recent research has shown the potential of utilizing different technologies, including cameras and laser scanners, to generate regular progress documentation of the site in the form of images, videos, or 3D scans (Mengiste et al., 2022). Combining classical construction project management methods such as LPS and reality capture presents opportunities to link the digital representation of tasks with their corresponding physical appearances at the construction site. Defining tasks based on the actual visual evolution of the construction elements within the lookahead planning framework provides a more accurate, real-time representation of project status, aiding in promptly identifying and rectifying potential issues. This combined approach can enhance LPS by producing accurate actual site appearance-based lookahead planning. Moreover, it sets the stage for effective automation of lookahead planning and management in construction. Thus, it represents a more holistic and realistic approach to project management, bridging the gap between planned tasks and actual physical progress.

Recognizing the need for a more accurate, comprehensive and automated approach, this study proposes a method for automatic lookahead schedule development, utilizing prior visual construction information. The aim is to prepare a deep learning model to predict lookahead schedules based on features extracted from site data, such as 2D and 360 images and point clouds representing prior construction performance and other influencing factors, including space utilization.

The rest of this paper is structured as follows: Section 2 offers an overview of relevant concepts and literature, emphasizing studies on constraint planning, lookahead planning and automation techniques. Section 3 introduces and explains the proposed methodology, illustrating its application to a sample construction project site. Section 4 presents the results obtained from the demonstration of the proposed method on the sample construction project, while Section 5 outlines the conclusions and indicates potential future prospects.

## 2 Background

Lookahead planning and performance forecasting have been investigated and used in the construction industry for decades. Lookahead planning often hinges on effective constraint management. Moreover, the labor-intensive and manual nature of gathering and processing construction progress data can hinder the prompt formulation of lookahead plans, consequently influencing construction site outcomes. In light of this, this section reviews literature that proposes automation techniques for lookahead planning, construction constraint planning, and the automated identification of construction execution patterns.

### 2.1 Automated lookahead planning

Construction involves multiple stakeholders with their own goals working together to achieve a given scope, which makes the industry prone to delays and cost overruns. Lean Construction principles are well-known for maximizing work value and minimizing waste in the construction process (Ballard, 2020). At its core, it tries to eliminate non-value-adding activities and improve overall project outcomes by emphasizing the importance of continuous improvement and teamwork.

LPS (Ballard, 2020) aims to improve the workflow that reduces lead times, cost and other forms of waste through effective communication and coordination among involved stakeholders. There are several stages in LPS, starting from the highest stage, such as master scheduling and going down to weekly work plans. The final execution planners usually produce the weekly work plans from information produced during lookahead planning. Lookahead planning (Ballard, 1997) in the LPS is the stage where the work that needs to be executed in the near future is detailed and planned after identifying potential constraints and possible mitigation plans (Seppänen et al., 2010).

Since its introduction, LPS has been implemented successfully in several countries (Bhatla et al., 2016; Murguia, 2019). The LPS, however, has its own limitations. One of the major challenges in the implementation of the LPS is that it depends on accurate information generation and collaboration (Perez and Ghosh, 2018). A case study by Perez and Ghosh (2018) reported barriers to adopting LPS. In their study, they identified that lookahead planning was not utilized enough due to a lack of complete and timely information on tasks and constraints from stockholders.

To address this limitation, automating lookahead planning has been of particular interest in recent years, with the advent of machine learning techniques. From rule-based systems (Darwiche et al., 1988; Desgagné-Lebeuf et al., 2019; Dong et al., 2013) and deep learning models (Alikhani et al., 2020; Amer et al., 2023, 2021b, 2021a), various approaches have been proposed and tested. Dong et al., (2013) developed a method of automating the generation of lookahead planning for finishing tasks by automatically converting information instances such as crew assignments, material, start and end dates and more.

Deep-learning-based schedule development and automation have been popular research topics in recent years. Amer and Golparvar-Fard (2019a, 2019b) developed a method to completely automate the generation of lookahead planning from manually annotated large construction activity data from historical construction projects of the same company. Their methods took advantage of the sequential data training capability of Long Short-Term Memory (LSTM)-based RNN, which is trained on construction sequences presented in terms of building construction vocabulary vectors. In their method, resource-based constraints and dependencies have been embedded in the activity names. However, these methods did not account for the semantic information from the actual construction site. Accounting for the actual site events, (Pal et al., 2024) proposed a method of estimating activity level percentage completion using semantic data such as images and point counts and an input 4D BIM. In their method, they performed segmentation of the construction work in progress from the point clouds using 2D semantic segmentation and projecting the segmentation masks to the 3D point cloud. The status of the construction work and the work completion percentage are estimated by comparing the ground truth masking of the elements from the 4D BIM with the segmented elements of the 3D point clouds. Although their method is promising in estimating the percentage of tasks completed and synthesizing site data with the project control systems, it still lacks estimates of the lookahead.

Similarly, Alikhani et al. (2020) proposed a bidirectional LSTM-RNN-based approach of automatic construction task sequencing from historical as-built schedule data. In terms of registering key knowledge from construction schedules to produce meaningful resources and train a deep learning algorithm, Zhao et al. (2020) proposed Natural Language Processing (NLP) and Ontology-based models, particularly for the application of automatic quality assessment of construction schedules. NLP was further used to develop a method of updating lookahead plans through the recognition of task dependencies (Amer et al., 2023).

There are several time series prediction models, such as autoregressive integrated moving average (ARIMA), exponentially weighted moving average (EMWA), or Holt-Winters; however, as an example, this study utilizes RNN. The model generally comprises five layers, where the input layer of the GRU RNN is the first point where the neural network obtains the data. This layer receives and shapes the information in every instance to produce input objects suitable for the hidden layer. It receives these timestep sequences and feeds them into the first Bidirectional GRU layer. This layer processes the timesteps and transforms them into a 64-dimensional feature space. The resulting output is then passed to the RepeatVector layer, duplicating the feature vector for each timestep, creating an appropriate input for the second Bidirectional GRU layer.

Automation methods are generally heavily linked with the scheduling templates, and the automated connection with the actual construction site is limited. In most cases, the actual information from the construction site is transferred to the automated scheduling method manually. Moreover, the constraints considered are limited to task dependencies and sometimes to the availability of materials. However, construction projects have several other constraints that are not straightforward to measure. In this regard, the following section explains one factor that affects the project performance but is often overlooked: space utilization.

## 2.2 Space utilization in construction lookahead planning

Extracting accurate and on-time information regarding the state of the construction site in terms of the special constraints requires special data processing to relate the material volume, built volume, and available space for movement and storage. Zhang and Cui (2021) proposed a genetic algorithm-based optimization solution to plan project space, coordinating the project schedule and material ordering. In addition to material storage, construction workspace monitoring is an important constraint when planning construction. In this regard, Messi et al. (2022) proposed a spatial conflict simulator using a game engine and Bayesian networks. However, these methods are proposed to simulate and produce accurate solutions during planning; to produce reliable lookahead planning, understanding available space and estimating the impact of space utilization during the construction activity on the upcoming construction activities is overlooked.

Several approaches have been proposed to estimate, model and plan construction space requirements to generate, allocate and resolve conflict (Hawarneh et al., 2021). Although it is generally understood that space and effective space allocation play a vital role in the performance of a construction project, most of the space allocation planning is not optimized with the course of the construction execution. The approaches proposed are commonly related to planning and are based on as-planned schedules, BIM (Tao et al., 2022) and logistics plans. Therefore, the performance of the space utilization plans can be affected by the level of detail of the BIM or the schedule. According to (Kolarić et al., 2022), improving the BIM LOD from LOD300 to LOD400 increases the performance of their proposed site logistic planning based on the site layout dynamic plan. Short-duration activities that utilize large and bulky construction equipment could be overlooked in project planning (Igwe et al., 2022), affecting work productivity in the operation stage. Therefore, it is important to consider the actual pattern of space utilization in the construction site during the lookahead planning to accurately represent constraints.

## 2.3 Appearance-based activity definition

Various devices and sensors have been employed to monitor activities at a construction site, and there are multiple ways to automatically convert raw data into a schedule of construction progress (Ekanayake et al., 2021). The use of sensing tools, such as RGB photo and video cameras, laser scanners, and 360-degree cameras, has been explored in literature. The most recent works, such as Deng et al. (2020), Martinez et al. (2021), Mengiste and García de Soto (2018), and others outlined in Alaloul et al. (2021), have introduced methods centered on material recognition, connecting material identification with project timelines to pinpoint potential delays. These studies predominantly lean on shapes, colors, or semantic CNN features as material defining attributes. However, the complexity of construction progress and material appearance makes it difficult to be wholly captured by these features alone. Some tasks with feature variations exceed the sensing capabilities of shapes and colors.

Furthermore, supervised learning models, such as those that depend on CNN features-based detection mechanisms, require extensive datasets to train machine learning models and effectively understand the textural variations between the spectrum of surface characteristics that could result in the progress of the construction work. Labeling visual information, especially to mirror specific construction statuses or material conditions, demands specialized training and expertise.

To resolve the limitations in feature definition and data size, material conditions can be recognized using specially engineered features or feature maps developed by combining commonly used features, such as CNN and texture maps, LBP, or GLCM (Mengiste et al., 2024). These appearance conditions require a close-up look into the patterns of the materials to differentiate them from a subsequent material condition that can be obtained after work is done. An example of this type of change in material condition is first- and second-coat plastering. The two layers of plastering have differences in texture, while they appear similar in color. This depiction of construction activity has been utilized in developing an as-built 4D BIM (Mengiste et al., 2023).

The present study utilizes material conditions to represent the state of the construction site by developing lookahead plans. The key motivation behind this approach is automating the data collecting and processing steps and closing the potential disconnect between the standard construction schedule and the actual site conditions. In certain scenarios, there may be a clear correlation between tasks in the standard schedule and the actual material conditions on the site. On the other hand, there could be instances where the standard schedule merges multiple states of appearance into one task. This consolidation can be challenging, as different workers may each contribute to individual material conditions. As a result, combining these could lead to misinterpretation. Therefore, construction material conditions offer a more accurate representation of efforts on a specific construction element.

## 3 Methodology

Our proposed approach represents the construction site activities regarding the appearance conditions of materials and space utilization. The actual work execution patterns are extracted in terms of material conditions and space utilization. These patterns are represented by features extracted from the prior construction activity using datasets such as Images and point clouds.

The visual data collected passes through a series of pre-processing steps to develop features before it is used for model development (Figure 1). These feature development processes include recognizing and extracting material conditions and computation of construction site spatial utilization information from point clouds and image data. The list of features and their corresponding descriptions with data inputs and outputs are presented in Table 1.

*Table 1: Feature development*

| Sub-section | Feature name | Feature Description | Input | Process output |
|---|---|---|---|---|
| 3.1.1 | Sequential material conditions | Sequential information on the appearance of construction materials representing the evolution of the construction site during the task execution. The representation of materials with the additional meta information regarding the time of the data collection. The time of data collection. | Images, meta information regarding the time of the data collection | Sequential text, name given to material conditions |
|  | percentage completion | Percentage of the measure of materials conditions appeared in the construction site during the data collection in reference to the total expected amount. | 4D BIM, Images | Percentage completion |
| 3.1.2 | Closeness | The proximity of the space-occupying objects to the building element under construction. This is measured by determining the location of the geometric center of the space-occupying element. | 4D BIM, Point clouds | x, y, z values |
|  | Utilization Extent | The amount of space taken by space-occupying elements around the construction element. This is measured as the ratio of occupied space to the total space around the element under construction. | 4D BIM, Point clouds | Numeric value between 0 - 1 |

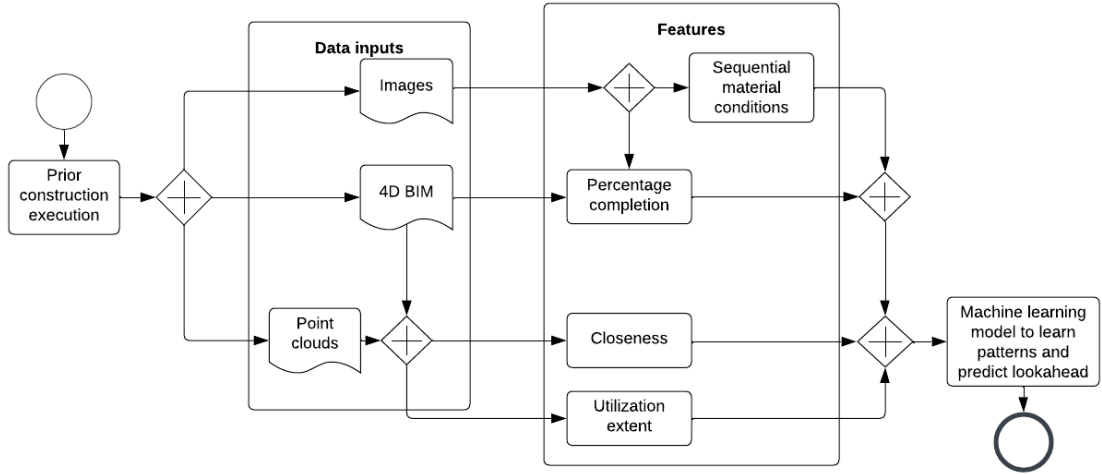

*Figure 1: Overall view of the proposed approach*

The implementation of the method proposed in this paper is composed of three major components. As shown in Figure 2, the first component is data collection from the construction site. Following the data collection, material condition information and construction site space utilization are extracted, processed and presented in tabular form. The assembled tabular data that resulted from the second part of the methodology is used to develop and evaluate the GRU RNN model.

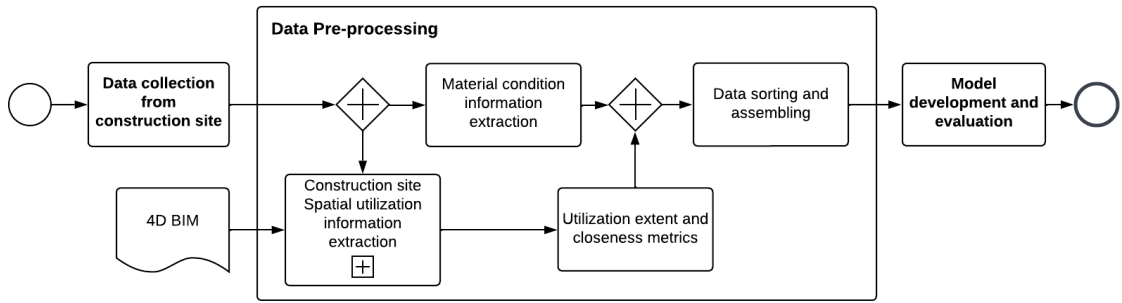

*Figure 2: Implementation process*

## 3.1 Data collection from construction site

In this study, visual information is represented in various formats, such as 2D RGB images, RGBD images, 360-degree panorama or colored point clouds. Therefore, data can be collected in any visual format. The construction site undergoes changes and progress every manhour and every day. The data collection, therefore, should preferably be as frequent as possible to be representative of the activities and the performance.

To illustrate the data processing procedure, a small construction site was chosen. Data was gathered both to develop the model and to verify its performance. The sample construction site is located at the NYUAD campus in Abu Dhabi, United Arab Emirates. The scope of the construction project was finishing work for new offices and labs, which included the construction of concrete masonry walls, finishing works such as plastering and painting, installation of electrical and air-conditioning elements, including breakers, junction boxes, switch sockets, air-conditioning ducts and hangers. Construction materials were stored in the construction space because the work occurred in a functioning building. For this study, it is assumed that the skill level of each construction worker stayed consistent throughout the project.

The sample construction project consisted of multiple rooms used as laboratory and office spaces. To illustrate the proposed methodology, we considered a single wall in an enclosed room, and the activities were monitored for 167 days. During that time, several data collection devices, such as 360 cameras and laser scanners (RGB and point cloud), were used to collect visual information. The devices used for the data collection are presented in Table 2. The work on the wall at the time of data collection involved finishing (plastering and painting) and electrical works (drilling for conduits, electrical fixture installation, such as distribution breakers (DBs), sockets and light switches).

The purpose of collecting 360-degree panorama images from the construction site was to process and recognize material conditions from textural features of material surfaces and to have a holistic representation of the space. Although the environment can be reconstructed from images, in this study, a separate laser scan was performed to capture the volumetric appearance of the space and to compute the space utilization pattern in the construction site. The 360-degree panorama images were collected every day except holidays and weekends, starting from February 02, 2022, till the completion of the construction on July 18, 2022. The laser scan data was collected on selected days.

*Table 2: Details of data collection and devices used*

| Data type | Dates | Time of data collection during the day | Devices used | Device specifications |
|---|---|---|---|---|
| Laser scan | 01/02/2022 12/02/2022 25/03/2022 15/04/2022 22/04/2022 29/06/2022 13/07/2022 | Noon to 1 pm | BLK360 | 360,000pts/sec 360 degrees horizontal and 300 vertical (Leica, 2019) |
| Panorama 360 | 02/02/2022- 18/07/2022 | Noon to 1 pm | GoPro Max | 5760 x 2880, image with 18MP quality (GoPro, 2023) |

Data representing the work execution plan and the actual timeline of the work was collected to evaluate the performance with the ground truth. Accordingly, the planned schedule indicated that concrete masonry work would be completed in 19 days, starting on Jan 1, 2022. The work included constructing a 37.6 square meter area of CMU work. After completing this task, work on two different tasks, Electrical works and Finishing (plastering and painting), were planned to start. Electrical work had three sub-tasks. The first sub-task was the installation of conduits and trays. This sub-task included the installation of 44 m of conduit, starting from March 19, 2022, and lasting 37 days. The second sub-task was wiring and cabling. On this subtask, 44 m of wiring and cabling work was expected to be completed in 54 days, starting April 01, 2022. The last sub-task in the Electrical work was the installation of DBs and fixtures. This task included the installation of one DB, two sockets, and one switch. According to the construction execution plan, finishing and plastering work started in parallel with electrical works. There were two sub-tasks under finishing works. The 37.6 square meters of surface was plastered and painted in 45 and 30 days, starting March 01 and April 15, 2022, respectively.

The complete as-built timeline showing the actual work completion on the wall of the area selected for this study is shown in Figure 3. The conduit installation and Wall drilling for conduit material conditions reached their completion around Feb 14, while CMU was completed before Feb 2. Both the first coat and second coat plastering works were started before Feb 5 and saw significant progress until Feb 20, when both works were completed by more than 80%. The plastering work remained static until May 30, and the remaining portion of the plastering was completed before Jun 2. The Epoxy painting work remained unfinished till the end of the period under consideration, when it started around March 7, and it was 71% completed by Jul 17. On the other hand, Fixing DBs and fixtures started before Feb 5 and progressed significantly for 15 days till Feb 20 and between Mar 22 and April 21. By the end of the data collection date, Fixing DBs and fixtures achieved 97% completion.

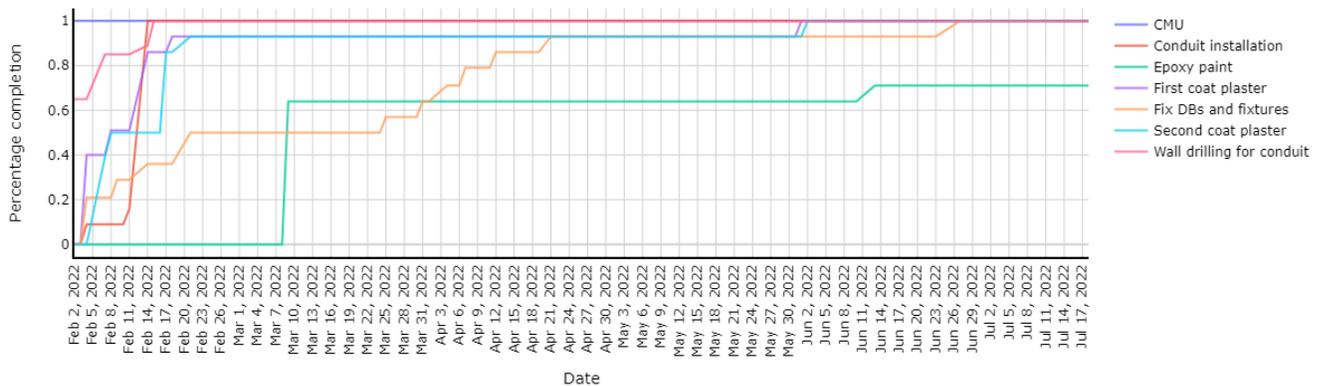

*Figure 3: Timeline showing actual project execution (% completion vs. time) for different materials in the space used for this study*

### 3.1.1 Material conditions and percentage completion

The collected data was initially processed to define and recognize the list of material conditions from the construction site. The extent of material conditions in the construction site is quantified, and the progress is evaluated based on the expected extent of material conditions at the time of data collection according to the master plan.

The sample images for different materials shown in Figure 5 were processed using methods proposed by Mengiste et al. (2022a) to detect corresponding material conditions.

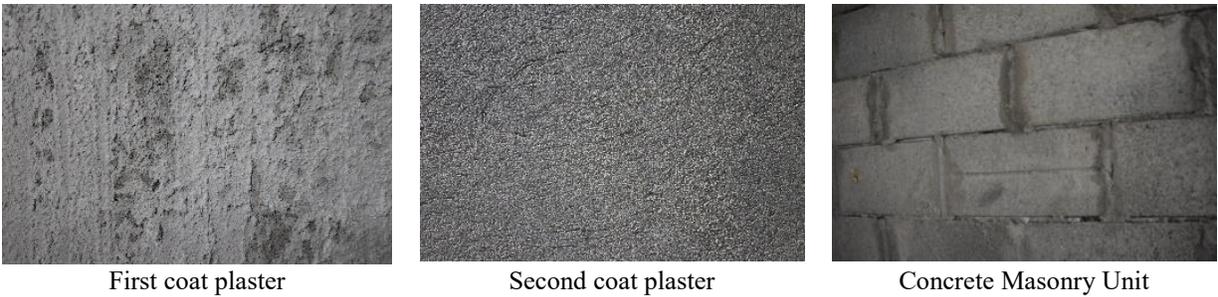

| First coat plaster | Second coat plaster | Concrete Masonry Unit |

*Figure 4: Sample images used to detect different material conditions*

The detected material condition can be integrated with 4D BIM (Mengiste et al., 2023) to compute the extent and quantities of the material conditions. Material condition recognition and quantity computation are beyond the scope of this study. However, quantity computing can be performed using Mengiste et al. (2022b) and Mengiste et al. (2023), where a special combination of texture features and descriptors were combined to define patterns and appearances of material conditions. These combinations further develop a machine-learning model to detect material conditions.

The percentage completion of the material conditions was quantified by comparing the size of the recognized material conditions with the as-planned BIM after integration. An example set of pictures displaying the change of material condition through time is presented in Figure 5.

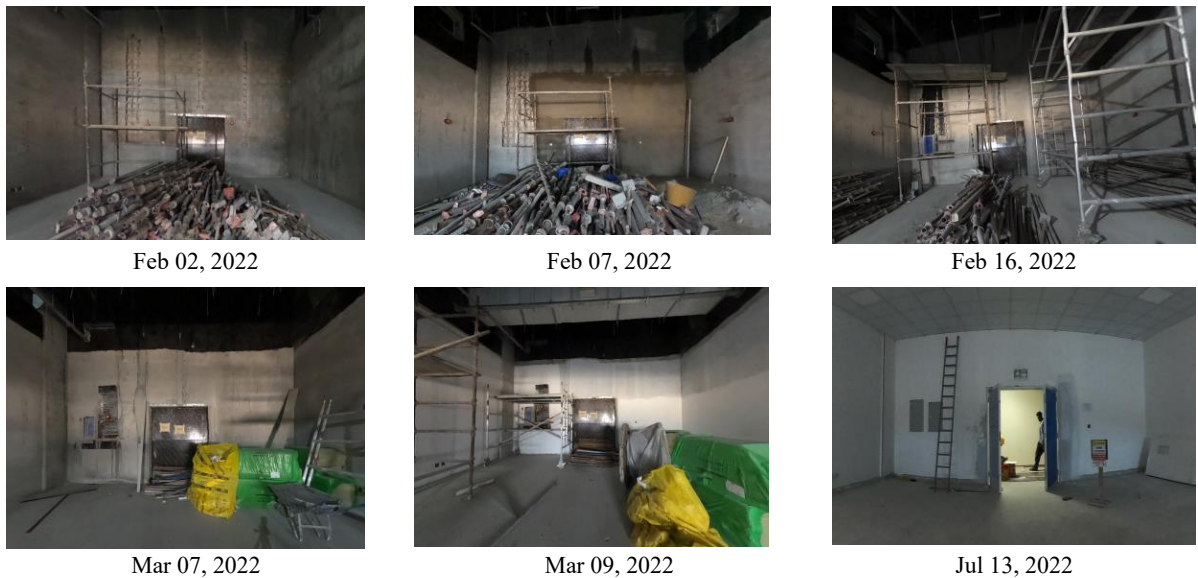

| Feb 02, 2022 | Feb 07, 2022 | Feb 16, 2022 |
| Mar 07, 2022 | Mar 09, 2022 | Jul 13, 2022 |

*Figure 5: Sample set of pictures showing the state of the selected space at different times during the project*

## 3.1.2 Spatial utilization

Spatial utilization of the construction site plays a crucial role in the execution performance of the construction. Therefore, the extent of free space within the site is evaluated based on the quantities of construction elements affected by temporary construction elements, such as scaffolding, materials stored or equipment used in an active construction space. In this study, spatial utilization is computed from laser scan data using the steps shown in Figure *6*.

*Step 1:* The point cloud is initially transformed into the predefined coordinate system in the planned 4D BIM.

*Step 2*: Vertexes are extracted from the corner points of the building elements that appeared in the planned 4D BIM at the specific time of the data collection.

*Step 3*: A set of building element enclosures was predefined from the extracted vertex points of the planned 4D BIM. These enclosures are volumetric geometries that are constructed by connecting vertexes. Considering the quality of the point cloud and error intensity, the enclosures can be set to have allowances over the vertexes to include points that are located with an offset.

*Step 4:* The transformed point cloud was registered to the enclosed volumes. Figure 7 presents the registration of a sample point cloud into enclosures generated from the BIM. Figure 7(a) shows the raw point cloud, and Figure 7(b) presents the corresponding point clouds after the registration of different enclosures representing different building elements.

*Step 5:* All points in the point cloud inside the predefined element enclosure are taken as points on the building. All remaining points, i.e., points outside the predefined enclosure, are taken as temporary elements occupying space. For instance, the purple points in Figure 7(b) were the temporary elements occupying the space, and the remaining elements, represented in red, green, lawn green, blue and dark blue points, represent the construction elements.

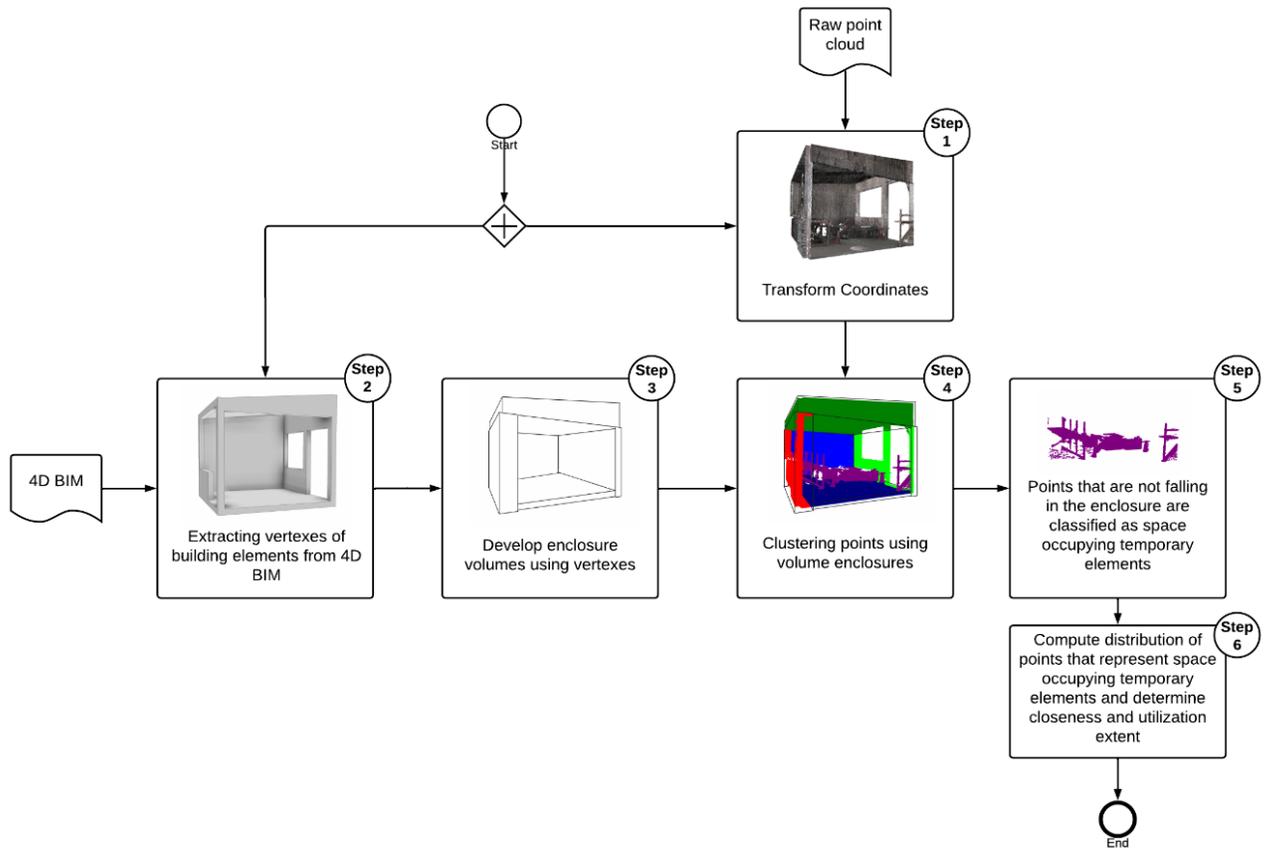

*Figure 6: Steps (1 through 6) used to extract information related to site spatial utilization.*

*Step 6*: Measuring the proximity of the accumulation of occupying objects to the target construction was conducted using the x, y and z components of the point cloud that are categorized as temporary and occupying space. Statistical analysis is performed to determine the concentration and localization of the temporary space-occupying objects relative to the total available space and the proximity of the construction element under consideration (e.g., Figure 8).

Figure 8(a) shows a histogram of the point clouds that represent the temporary and occupying objects. Figure 8(b) shows the statistical distribution of the remaining points on the floor. The relevant factors for this study are generalized in terms of Utilization Extent and Closeness.

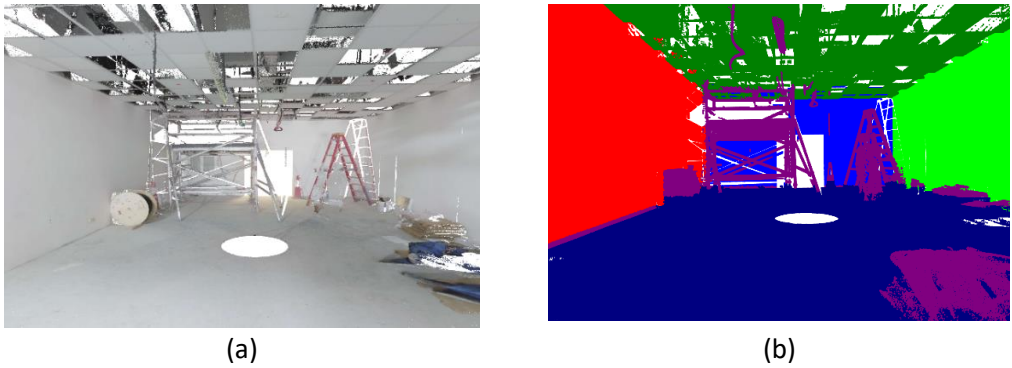

(a)             (b)

*Figure 7: Sample point cloud pre-processing showing (a) raw point cloud and (b) point cloud after points are classified into corresponding permanent building element enclosures.*

Utilization Extent is used to measure the occupation of the space around the construction site in general. In this study, the Utilization Extent is defined based on the number of points that represent temporarily occupied points in reference to the number of points that represent temporary objects and the floor itself. This parameter ranges from 0 to 1; it is zero when the number of points designated as temporary space-occupying elements is relatively insignificant compared to the number of points representing the floor. On the other hand, the parameter becomes close to 1 when temporary objects completely occupy the space, and the number of points representing these objects is relatively large, while the number of points representing the floor is insignificant. The number of points on the floor could be insignificant for two main reasons. The first is where the floor is completely covered with temporary objects, and the second is when the objects placed in the room are large enough to cover the floor, resulting in occlusion.

Closeness measures the proximity of the space occupying elements to the building element under construction. This parameter is calculated as a spatial centroid of the space-occupying temporary elements with a static reference point; in this case, the mean x, y and z values of the temporary and space-occupying point cloud (Figure 8(a)) that represent the Closeness of the temporary objects to the origin of the predefined coordinate system.

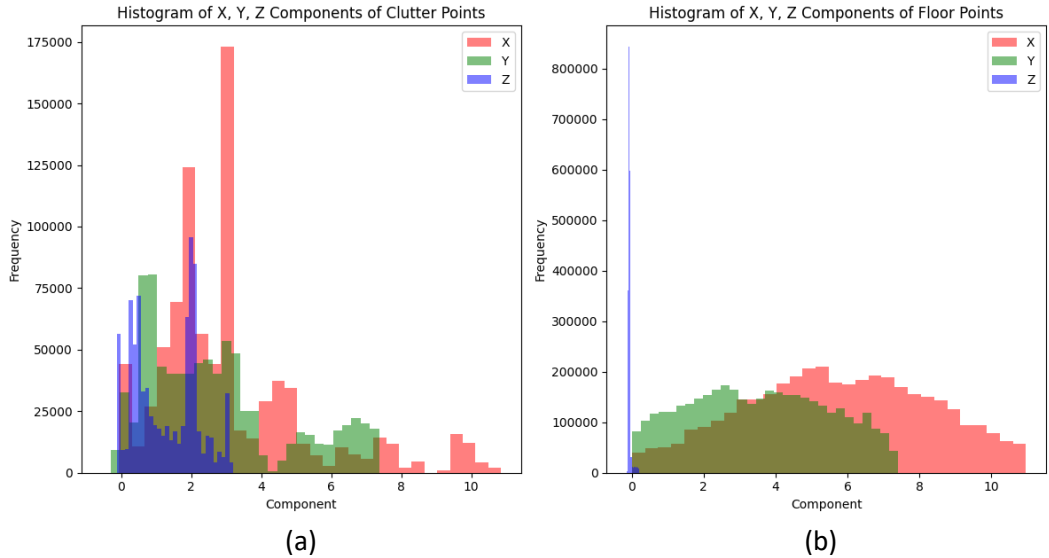

*Figure 8: Sample statistical analysis of point cloud for spatial utilization. (a) Point cloud representing temporary and occupying objects. (b) Points on the floor*

A subset of the feature dataset developed for the sample project using the preprocessing methods outlined in this study is presented in Table 3, which contains the date of data collection, percentage execution of the work as defined by the appearance of the given material conditions, and the spatial utilization information in terms of closeness of the bulk temporary elements in the space and the Utilization Extent. The data represents every day during the project timeline except for the non-working periods and holidays from February 2 to July 18, 2022.

Table 3: Sample dataset

| Date | Percentage execution | Material conditions | Closeness | | | Utilization Extent |
|---|---|---|---|---|---|---|
| | | | x | y | z | |
| 2-Feb-22 | 100% | CMU | 4.39 | 3.77 | 0.82 | 0.44 |
| 7-Jun-22 | 100% | CMU | 5.64 | 3.31 | 0.48 | 0.26 |
| 8-Jun-22 | 100% | CMU | 5.64 | 3.31 | 0.48 | 0.26 |
| : | : | : | : | : | : | : |
| 2-Feb-22 | 0% | First coat plaster | 4.39 | 3.77 | 0.82 | 0.44 |
| 7-Jun-22 | 100% | First coat plaster | 5.64 | 3.31 | 0.48 | 0.26 |
| 8-Jun-22 | 100% | First coat plaster | 5.64 | 3.31 | 0.48 | 0.26 |
| : | : | : | : | : | : | : |
| 2-Feb-22 | 0% | Second coat plaster | 4.39 | 3.77 | 0.82 | 0.44 |
| 3-Feb-22 | 0% | Second coat plaster | 4.39 | 3.77 | 0.82 | 0.44 |
| 8-Jun-22 | 100% | Second coat plaster | 5.64 | 3.31 | 0.48 | 0.26 |
| : | : | : | : | : | : | : |
| 2-Feb-22 | 0% | Epoxy paint | 4.39 | 3.77 | 0.82 | 0.44 |
| 3-Feb-22 | 0% | Epoxy paint | 4.39 | 3.77 | 0.82 | 0.44 |
| 8-Jun-22 | 64% | Epoxy paint | 5.64 | 3.31 | 0.48 | 0.26 |
| : | : | : | : | : | : | : |
| 2-Feb-22 | 0% | HVAC_paint | 4.39 | 3.77 | 0.82 | 0.44 |
| 3-Feb-22 | 0% | HVAC_paint | 4.39 | 3.77 | 0.82 | 0.44 |
| 8-Jun-22 | 29% | HVAC_paint | 5.64 | 3.31 | 0.48 | 0.26 |
| : | : | : | : | : | : | : |
| 2-Feb-22 | 65% | Wall drilling for conduit | 4.39 | 3.77 | 0.82 | 0.44 |
| 3-Feb-22 | 65% | Wall drilling for conduit | 4.39 | 3.77 | 0.82 | 0.44 |
| 8-Jun-22 | 100% | Wall drilling for conduit | 5.64 | 3.31 | 0.48 | 0.26 |
| : | : | : | : | : | : | : |
| 2-Feb-22 | 0% | Conduit installation | 4.39 | 3.77 | 0.82 | 0.44 |
| 3-Feb-22 | 0% | Conduit installation | 4.39 | 3.77 | 0.82 | 0.44 |
| 8-Jun-22 | 100% | Conduit installation | 5.64 | 3.31 | 0.48 | 0.26 |
| : | : | : | : | : | : | : |
| 2-Feb-22 | 0% | Fix DBs and fixtures | 4.39 | 3.77 | 0.82 | 0.44 |
| 3-Feb-22 | 0% | Fix DBs and fixtures | 4.39 | 3.77 | 0.82 | 0.44 |
| 8-Jun-22 | 93% | Fix DBs and fixtures | 5.64 | 3.31 | 0.48 | 0.26 |
| : | : | : | : | : | : | : |

### 3.2 Model development and evaluation

To implement the proposed approach, GRU RNN was trained using the developed set of features, including material conditions and spatial utilization information to estimate the percentage completion of tasks in the coming weeks. GRU RNN was selected as an example to take advantage of its simpler architecture and, potentially, a small number of parameters. Reducing parameters decreases the chance of overfitting; this is particularly important when the number of data available for training is limited.

As shown in Table 3, raw data is comprised of textual entries (Material Conditions), dates (Data Collection Date), and numeric values (the remaining special information and metrics). Before feeding this data into the deep learning model, it undergoes a series of pre-training steps. Textual data (Material Conditions) is converted into machine-readable format through Binary Encoding. The date information is decomposed into three distinct attributes: Date, Month, and Year. The remaining numeric information is scaled using the MinMaxScaler (Pedregosa et al., 2011) to bring all values into a consistent range. This processed dataset, including the encoded values, reshaped date information, and scaled numeric data, is then segmented into sequences of 18 timesteps each.

The Hidden layer contains two layers of bidirectional GRU and a report vector layer: it processes and reshapes values to produce predictions. The first hidden bidirectional GRU layer captures dependencies in both directions and simultaneously uses hyperbolic tangent and sigmoid activation functions in hidden and recurrent states. This layer generally does not produce predictions but generates encoded information to be decoded by the second bidirectional GRU layer. The repeat vector is important in the model because it replicates the encoded information generated in the first bidirectional GRU to match the input shape of the second bidirectional GRU. The second bidirectional GRU differs from the first one because it produces an output sequence in the form of prediction from every batch of processed data. The final layer is the Output layer, in which the sequence of predictions produced in the last hidden layer is transposed linearly. This layer has a single output unit for every timestep, where the activation function is set to be linear to allow the model to produce a continuous prediction.

The model was developed using *Keras* (Chollet, 2015) on a TensorFlow platform (Abadi et al., 2015). The total number of trainable parameters created in the model development was 104,577 of which around 29,952 were created on the first bidirectional hidden layer, and 74,496 parameters were created in the second hidden bidirectional GRU layer. The last dense layer produced 129 parameters while training the model.

In this study, an 80:20 split was used to partition the dataset for training and testing purposes. Out of the total 56 data points, the last 18 data instances (20% of the total data) were manually selected to form the testing set. The training dataset was further subdivided to facilitate model development and evaluation. Specifically, a portion of the training data, constituting 10%, was reserved for validation, while the remaining 90% was utilized for the actual model training. These last 18 data instances, chosen to evaluate lookahead performance, represent a continuous sequence of the last 18 days in the dataset, with each data instance corresponding to a single day. Once the model was trained and validated, it was then employed to estimate the performance of the construction process during these last 18 days. The reported performance metrics of the model are based on its performance during this period, which was intentionally hidden from the model during both the training and validation phases.

The model estimates the percentage completion of multiple activities on the construction site. The model evaluation considered the Mean Squared Error (MSE) and Mean Absolute Error (MAE) to assess normally distributed errors. The MAE measures the average magnitude of the errors in a set of predictions without considering their direction; thus, it is a linear score, which means all the individual differences are equally weighted. On MSE, errors are squared before they are averaged: MAE shows the performance of the model in smaller errors, while MSE is useful to visualize the magnitude of offset when the errors are larger.

On every iteration, the model was evaluated using MSE and MAE. As shown in Figure 9(a), after the 40[th] epoch, the model consistently had less than 15% MAE during validation and less than 5% during training after the 80[th] epoch. Similarly, the model resulted in less than 3% MSE on validation data and less than 1% in training data (Figure 9(b)). The model was tested using a dataset not used during training and validation; using the testing data, the model's overall performance was 16.57% and 5.82% for MSE and MAE, respectively. Given the size of the data, these results show that the model is not overfitting.

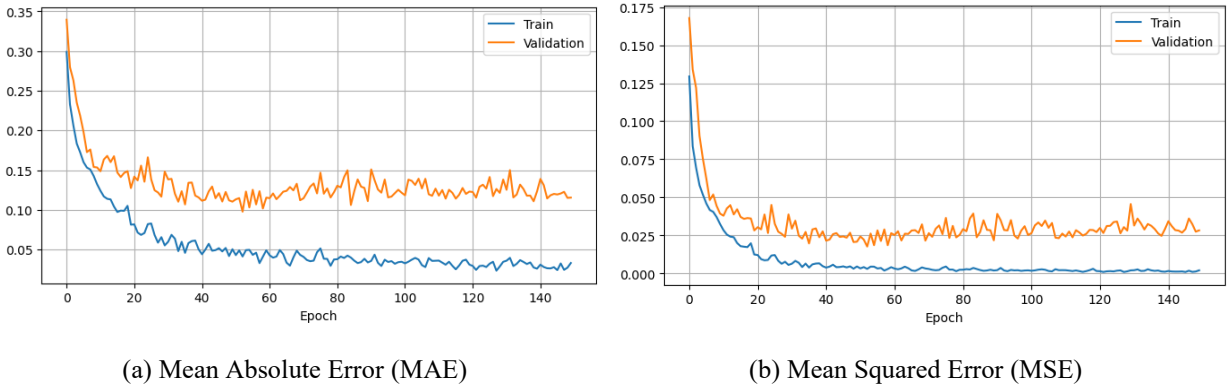

(a) Mean Absolute Error (MAE)          (b) Mean Squared Error (MSE)

*Figure 9: Model performance*

This study implemented a structured approach to hyperparameter optimization, aiming to fine-tune the model's performance. A grid search mechanism over a predefined parameter space was used to define the maximum result yielding GRU units and learning rate values. For each combination, the model was trained on a consistent dataset, ensuring uniformity for comparison. The primary metrics for evaluation were the MAE and the MSE on the validation set.

The results revealed significant variations in performance based on hyperparameter settings. Certain combinations led to convergence on lower MAE and MSE values during training and validation. Lower MAE increases the confidence of predicted completion and narrows down the uncertainty values in estimation, while decreased MSE values reduce the occurrences of errors and outliers during prediction. The combination of a 0.001 learning rate and GRU Unit values of 64 provided a maximum performance of the model in terms of both MAE and MSE.

## 4 Results and discussion

The model was used to produce a performance forecast of construction activities for the last 18 days (from June 10 to July 18, 2022). The actual execution information of these days was used as a ground truth. Until June 9th, most of the tasks were finalized, excluding those related to Epoxy paints and fixing DBs and fixtures material conditions (Figure 3). Consequently, the performance forecast is exclusively presented (Table 4) for these two outstanding tasks to provide a visual comparison between the forecasted results and the actual performance data.

In practical terms, developing an accurate lookahead plan is highly dependent on understanding and managing the constraints that could arise from the involvement of several stakeholders and a complex construction site environment. Automating the process of reading past patterns in relation to significant site-specific or location-specific constraints, such as space utilization, could quickly help manage the construction site, produce remedies for potential risks and boost productivity. Anchoring the lookahead plan with the site appearance characteristics could help understand the construction site and produce more visible plans.

Lookahead planning is related to productivity; the results from our demonstration can be used to determine the expected work productivity. The MAE provides insight into the average deviation one might expect when predicting daily performance. Essentially, this daily performance prediction can translate to a daily productivity expectation. While MAE and MSE offer a measure of prediction accuracy, their sensitivity to errors differs. The MSE, due to its squaring of errors, is particularly sensitive to large deviations. Therefore, a high MSE relative to MAE might suggest the existence of larger errors or potential outliers. Conversely, a lower MSE suggests that the errors the model produced are more uniformly distributed and that larger errors are less frequent.

Considering the performance of the model (5.82% MSE and 16.57% MAE), we can conclude the following: for tasks under evaluation, the performance prediction carries an average error of approximately 16.57%. In practical terms, the performance on July 18 (Table 4) of the task epoxy painting is projected to be 79% complete. However, factoring in the MAE, the construction manager should prepare for a performance range between 62.43% (79% - 16.57%) and 95.57% (79% + 16.57%). The MSE value of 5.82%, being lower than the MAE, reinforces the notion that larger errors or extreme deviations out of this bound are less frequent in the model predictions. The actual execution lies between the median and the lower bound, and the occurrence of the upper limit results in a larger error. In the case of Fixing DBs and fixtures, the actual completion lies between the upper limit and the median estimation of the GRU RNN model.

*Table 4: GRU RNN model results*

| Task | Planned % complete | | Actual % complete | | Estimated % completion on July 18 (GRU RNN) | | |
|---|---|---|---|---|---|---|---|
| | June 9 | July 18 | June 9 | July 18 | Upper limit (+MAE) | Median | Lower limit (-MAE) |
| Fix DBs and fixtures | 100 | 100 | 93 | 97 | 100 | 94 | 77.43 |
| Epoxy painting | 100 | 100 | 64 | 71 | 95.57 | 79 | 62.43 |

In general, the GRU RNN-based model can forecast task completion rates. The error range that is shown from the MAE score is large. Although the results show that the approach is in the right direction, further work is required to reduce the error bounds and the MAE score. Effective site performance forecasting could enable optimal resource allocation and efficient planning. With this, delays can be predicted and mitigated early, drastically reducing costs and time overruns. The integration of factors such as space utilization in the model can also lead to improved workspace organization, promoting even better productivity. More so, by estimating non-working idle times, the model can highlight unproductive patterns, providing an opportunity for process improvement and further boosting productivity.

## 4.1 Limitations and future work

The practical application of the GRU RNN-based lookahead model has shown promising results in predicting task completion on a simple construction site. However, one of the primary limitations is that the model was demonstrated on a construction site that includes only finishing tasks. Moreover, the skill level of workers was assumed to be consistent throughout the project. The effectiveness of the model on larger, more complex construction sites remains untested.

Additionally, like many deep learning models, the performance of this GRU RNN-based model is heavily reliant on the size and quality of the training data. The model has been trained on a limited dataset, which might not cover all possible scenarios, activities, or task patterns found in a larger, more diverse construction environment. This was also visible in the performance metrics reported. For future work, it would be beneficial to test and refine the model using larger and more diverse datasets collected from various construction activities across different types of sites. This would help increase the robustness of the model, allowing it to capture a broader range of activity patterns and relationships. Moreover, the size of the data used affects the MAE and MSE scores obtained. In general, a larger data set would help reduce the error range and improve accuracy in estimating performance and productivity.

It could also be worth exploring the integration and influence of other relevant factors in addition to space utilization, such as worker skills, equipment availability, or material supply delays. This integration could help produce a comprehensive approach with enhanced predictive accuracy.

## 5   Conclusion

The automatic lookahead forecasting GRU RNN-based model predicts task completion on construction sites by considering various factors like prior work execution patterns represented in the form of material conditions and space utilization of the construction site. Space utilization was measured based on Utilization Extent and Closeness metrics, which are computed from the proximity and the area occupied by the non-construction objects around the construction element. The model showed results on a simple construction site with minimum overall MAE and MSE of 16.57% and 5.82%, respectively. The proposed method cuts down the time and effort of processing multi-source data from several stakeholders and provides promising results in managing constraints and producing reliable lookahead planning to help avoid project delays.

**Acknowledgment**

This work was supported by the NYUAD Center for Sand Hazards and Opportunities for Resilience, Energy, and Sustainability (SHORES). This work also benefited from the collaboration with the NYUAD Center for Interacting Urban Networks (CITIES), funded by Tamkeen under the NYUAD Research Institute Award CG001. The authors would like to thank William Fulton and Mohamed Rasheed from the Campus Planning and Projects Office at NYUAD for their support and for allowing access to ongoing projects on campus to be used for our research.